\documentclass[11pt]{article}

\usepackage[T1]{fontenc}
\usepackage[utf8]{inputenc}
\usepackage{lmodern}
\usepackage{amsmath,amssymb}
\usepackage{booktabs}
\usepackage{float}
\usepackage{graphicx}
\usepackage[hidelinks]{hyperref}

\ifdefined\pdfsuppresswarningpagegroup
\fi

\title{TEMPER: Temporal Encoder-Masked Probabilistic Ensemble Regressor for Time-Series Forecasting}

\author{
Giancarlo Vercellino\\
Independent researcher\\
\texttt{giancarlo.vercellino@gmail.com}
}

\date{September 1, 2026}

\begin{document}
\maketitle

\begin{abstract}
Probabilistic forecasting requires accurate central predictions and calibrated uncertainty estimates. This paper presents TEMPER, the Temporal Encoder-Masked Probabilistic Ensemble Regressor, a univariate time-series forecasting algorithm that combines a temporal autoencoder, a differentiable masked neural decision forest, continuous ranked probability score (CRPS) training, and Gaussian-mixture post-processing. The R implementation is built on \texttt{torch} for R and returns horizon-wise density, distribution, quantile, and sampler functions. We evaluate TEMPER on three deterministic synthetic level series with trend, periodic, regime-switching, nonlinear-threshold, and heteroskedastic components. Across 96 rolling-origin forecasts at horizons $t+1$, $t+5$, $t+20$, and $t+60$, TEMPER obtains 2.824\% mean CRPS normalized by origin level, 3.635\% median absolute error, and 68.8\% empirical 90\% interval coverage after training with a 300-epoch cap and early-stopping patience of 100. A naive persistence bootstrap has the best aggregate CRPS, 2.763\%, while TEMPER has the best median absolute error and the best CRPS at $t+1$ and $t+5$. The ablation study uses matched series-origin-horizon cells, horizon-wise CRPS deltas, endpoint sensitivity summaries, and a calibration-specific interval study. Relaxing the learned mask improves average CRPS by 0.472 percentage points on the ablation subset, mainly through long-horizon gains. A twofold interval inflation improves held-out coverage from 54.2\% to 91.7\% and gives the best 90\% interval score among tested calibration rules. The results identify calibration, horizon-specific tuning, and component selection as the central research priorities.
\end{abstract}


\section{Introduction}

Many forecasting workflows still report a single path even when downstream decisions depend on tail risk, scenario dispersion, and the probability of crossing operational thresholds. Predictive distributions address this problem by representing uncertainty directly. They also require different modeling and evaluation tools: a probabilistic forecaster should be calibrated, sharp, and robust under distribution shift rather than merely accurate in point error.

TEMPER, the Temporal Encoder-Masked Probabilistic Ensemble Regressor, is a hybrid architecture for univariate probabilistic forecasting. It reframes a level series into lagged relative changes, encodes each input window into a latent vector, feeds the latent state to a neural decision forest with differentiable stochastic feature masks, trains the model with an ensemble CRPS objective plus reconstruction regularization, and smooths the resulting forecast samples with horizon-specific Gaussian mixtures. The R package implementation is written around \texttt{torch} tensors and \texttt{nn\_module} components. The \texttt{torch} package provides PyTorch-like neural-network programming in R through the LibTorch backend \cite{torchcran2025,paszke2019pytorch}; TEMPER uses this stack for automatic differentiation, tensor operations, and optimizers.

This paper makes four contributions.

\begin{itemize}
  \item It gives a formal presentation of the TEMPER algorithm and its training objective.
  \item It documents the reference R/\texttt{torch} implementation's data transformation, neural forest, mixture smoothing, and fan-chart outputs.
  \item It evaluates the method in a rolling-origin probabilistic forecast study on synthetic series with known temporal structure, using naive persistence, random-walk, and AR(1) bootstrap baselines.
  \item It presents a paired ablation design with horizon-wise deltas, endpoint sensitivity summaries, and a calibration-specific interval study.
\end{itemize}

The paper is framed for statistical machine learning rather than as an application-specific forecasting claim. The focus is estimation with a proper scoring rule, differentiable ensemble structure, stochastic feature masking, representation learning, distributional evaluation, and ablation-based model analysis. The synthetic design gives a controlled setting in which the data contain learnable temporal structure, while still making it possible for simple baselines to remain competitive.

\section{Related Work}

TEMPER combines ideas from probabilistic forecast evaluation, representation learning, differentiable tree ensembles, stochastic relaxations, and conformal calibration. Proper scoring rules provide the evaluation and estimation foundation for distributional forecasts; the CRPS is especially attractive for real-valued outcomes because it generalizes absolute error to predictive distributions \cite{matheson1976scoring,gneiting2007proper,gneiting2007calibration}. CRPS-based learning has also been used for adaptive aggregation of probabilistic forecasts \cite{berrisch2023crps}. TEMPER uses CRPS directly as the ensemble training objective.

Deep probabilistic forecasting has developed several influential architectures. DeepAR uses autoregressive recurrent networks to estimate predictive distributions across related time series \cite{salinas2020deepar}. Deep state-space models combine recurrent parameterization with classical state-space structure \cite{rangapuram2018deep}. Temporal Fusion Transformers add attention, gating, and feature selection for interpretable multi-horizon forecasting \cite{lim2021temporal}. N-BEATS and N-HiTS show that deep residual and hierarchical interpolation structures can be effective for univariate and long-horizon forecasting \cite{oreshkin2020nbeats,challu2023nhits}. Toolkits such as GluonTS have helped standardize probabilistic neural forecasting workflows and benchmarks \cite{alexandrov2020gluonts}.

TEMPER is closest in spirit to this neural forecasting literature but uses a different inductive bias. Rather than a recurrent, attention, or residual-block decoder, it forecasts through a differentiable decision-forest ensemble in a learned latent return space. Its distributional output is produced by a horizon-wise Gaussian-mixture smoother over ensemble paths, so the neural forecasting references are used as architectural context rather than as direct empirical baselines.

Neural autoencoders have long been used to learn compact representations of high-dimensional observations \cite{hinton2006reducing}. TEMPER uses the autoencoder as a temporal bottleneck: it compresses a window of lagged returns and regularizes the latent representation through reconstruction. Decision forests are attractive because they aggregate multiple partitioning rules, while neural decision forests make tree routing differentiable and trainable end to end \cite{breiman2001random,kontschieder2015deep}. TEMPER follows this differentiable-forest spirit but uses the trees as an ensemble of forecast generators rather than as a classifier. Its feature mask uses a Gumbel-Softmax-style continuous relaxation \cite{jang2017categorical,maddison2017concrete}, allowing stochastic feature selection to participate in gradient-based training.

The calibration analysis is related to conformal prediction, which wraps predictive algorithms with finite-sample coverage guarantees under exchangeability assumptions \cite{vovk2005algorithmic,angelopoulos2021gentle}. The small split-conformal exercise in this paper is not presented as a final solution for time-dependent coverage, but it provides a diagnostic contrast between raw sharp intervals, fixed-width inflation, and residual-based calibration.

\section{The TEMPER Algorithm}

Let $\{s_t\}_{t=1}^{T}$ be a univariate positive level series. TEMPER first converts levels to relative changes,
\begin{equation}
r_t = \frac{s_t}{s_{t-1}} - 1,\qquad t=2,\ldots,T,
\end{equation}
after imputing missing values when needed. For a lookback length $P$ and forecast horizon $H$, sliding windows form supervised pairs
\begin{equation}
x_i = (r_i,\ldots,r_{i+P-1}) \in \mathbb{R}^{P}, \qquad
y_i = (r_{i+P},\ldots,r_{i+P+H-1}) \in \mathbb{R}^{H}.
\end{equation}

\subsection{Temporal Encoding}

An autoencoder maps each input window to a latent vector:
\begin{equation}
z_i = f_{\theta}(x_i), \qquad \hat{x}_i = g_{\phi}(z_i),
\end{equation}
where $z_i \in \mathbb{R}^{d}$ and $d$ is the user-selected latent dimension. In the reference implementation, both encoder and decoder are feed-forward networks with one hidden layer and ReLU activation. The reconstruction term encourages the latent state to retain information about the input window:
\begin{equation}
\mathcal{L}_{\mathrm{rec}} = \frac{1}{N}\sum_{i=1}^{N}\lVert x_i-\hat{x}_i\rVert_2^2.
\end{equation}

\subsection{Masked Neural Decision Forest}

The latent vector is passed to an ensemble of $M$ soft decision trees. Each tree starts with a learnable mask. During training, latent feature $j$ is multiplied by
\begin{equation}
m_j = \sigma\left(\frac{\log \alpha_j + g_j}{\tau}\right),
\end{equation}
where $g_j$ is Gumbel noise, $\tau$ is the temperature, and $\alpha_j$ is parameterized by the initial keep probability. At evaluation time the implementation uses the deterministic sign of $\log \alpha_j$.

For a tree of depth $D$, there are $2^D-1$ internal nodes and $2^D$ leaves. Each internal node computes a soft routing probability
\begin{equation}
p_k(z) = \sigma(w_k^\top (m \odot z)+b_k).
\end{equation}
Leaf path probabilities are the products of the corresponding left and right routing probabilities. If $\pi_{\ell}(z)$ is the probability of reaching leaf $\ell$ and $v_{\ell}\in \mathbb{R}^{H}$ is the leaf forecast vector, tree $m$ emits
\begin{equation}
\hat{y}^{(m)}(z) = \sum_{\ell=1}^{2^D}\pi_{\ell}(z)v_{\ell}.
\end{equation}
Stacking all trees yields an ensemble $\{\hat{y}^{(m)}\}_{m=1}^{M}$ of multi-step return forecasts.

\subsection{CRPS Ensemble Loss}

For each horizon, TEMPER minimizes an empirical CRPS for ensemble samples. For an observed target $y$ and ensemble values $\hat{y}^{(1)},\ldots,\hat{y}^{(M)}$, the sample CRPS is
\begin{equation}
\widehat{\mathrm{CRPS}} =
\frac{1}{M}\sum_{m=1}^{M}\left|\hat{y}^{(m)}-y\right|
- \frac{1}{2M^2}\sum_{m=1}^{M}\sum_{m'=1}^{M}
\left|\hat{y}^{(m)}-\hat{y}^{(m')}\right|.
\end{equation}
The training objective averages this score over batches and horizons and adds reconstruction regularization:
\begin{equation}
\mathcal{L} = \mathcal{L}_{\mathrm{CRPS}} + \lambda_{\mathrm{rec}}\mathcal{L}_{\mathrm{rec}}.
\end{equation}
The experiments use Adam \cite{kingma2015adam}.

\subsection{Level-Space Predictive Distributions}

The neural forest forecasts relative changes. For a forecast origin $T$, each ensemble path is converted back to level space by
\begin{equation}
\hat{s}^{(m)}_{T+h} = s_T \prod_{j=1}^{h}\left(1+\hat{y}^{(m)}_j\right),
\qquad h=1,\ldots,H.
\end{equation}
For each horizon, the resulting finite sample is smoothed by a Gaussian mixture. The implementation fits candidate $k$-means partitions, chooses a component count by an elbow-style curvature heuristic, estimates component means, standard deviations, and weights, and returns density, distribution, quantile, and random-generation functions.

\subsection{Computation and Implementation}

For a batch of size $B$, latent dimension $d$, forecast horizon $H$, forest size $M$, and tree depth $D$, the main forest pass evaluates $M(2^D-1)$ internal routing nodes and aggregates $M2^D$ leaf predictions. The ensemble CRPS term adds a pairwise sample distance cost of order $O(BHM^2)$. This quadratic dependence on ensemble size is one reason that the ablation study includes a reduced-ensemble variant.

The implementation is written in R using \texttt{torch} tensors, modules, automatic differentiation, and the Adam optimizer. The experiments use single-threaded \texttt{torch} execution to reduce run-to-run timing variability on CPU. The package interface returns fitted forecast functions rather than only arrays, so downstream users can evaluate densities, probabilities, quantiles, or random samples at each horizon.

\section{Experiments}

\subsection{Synthetic Data}

The benchmark uses three deterministic synthetic positive level series of length 900. Each series is generated through clipped log returns
\begin{equation}
u_t = \mu_j(t,u_{t-1}) + \phi_j u_{t-1} + \epsilon_t,\qquad
\epsilon_t \sim \mathcal{N}(0,\sigma_j(t)^2),
\end{equation}
followed by $s_t=s_{t-1}\exp(u_t)$. The clipping range is $[-0.08,0.08]$ to avoid unrealistic explosive paths. The three series are:

\begin{itemize}
  \item \textbf{CycleTrend:} a low-noise trending process with 20-day and 60-day sinusoidal components.
  \item \textbf{RegimeCycle:} a cyclic process with alternating drift regimes and alternating volatility regimes.
  \item \textbf{ThresholdWave:} a nonlinear process with thresholded sinusoidal drift and mean reversion.
\end{itemize}

The synthetic date index runs from January~1, 2020 through June~18, 2022. The exact generator and random seeds are contained in the ancillary experiment script.

\subsection{Protocol}

We use a rolling-origin design with eight forecast origins: August~17, 2021; September~21, 2021; October~26, 2021; November~30, 2021; January~4, 2022; February~8, 2022; March~15, 2022; and April~19, 2022. At each origin and for each series, models produce 60-step forecasts. Metrics are reported at four selected horizons, $t+1$, $t+5$, $t+20$, and $t+60$. This gives $3\times 8\times 4=96$ evaluated forecast-horizon pairs per benchmark model.

TEMPER is trained with lookback $P=60$, forecast horizon $H=60$, latent dimension 10, 80 trees, depth 5, initial mask keep probability 0.8, mask temperature 0.5, maximum mixture components 8, training rate 0.75, batch size 16, Adam learning rate 0.003, reconstruction weight 0.3, 300 maximum epochs, and early-stopping patience 100. Each predictive distribution is evaluated with 300 samples.

We compare against three simple probabilistic baselines \cite{hyndman2021forecasting}: a naive persistence bootstrap, a random-walk bootstrap, and an AR(1) residual bootstrap. Metrics are normalized by the level at the forecast origin. We report mean CRPS, median absolute error (MdAE), empirical 90\% interval coverage, mean 90\% interval width, average wall-clock training or simulation time per series-origin fit, and the number of evaluated horizons.

\subsection{Benchmark Results}

\begin{table}[H]
\centering
\caption{Rolling-origin probabilistic forecast results across three synthetic series, eight origins, and four selected horizons. CRPS, MdAE, and interval width are percentages of the level at the forecast origin. Epochs is the mean actual number of training epochs after early stopping.}
\label{tab:benchmark}
\resizebox{\linewidth}{!}{\begin{tabular}{lrrrrrrr}
\toprule
Model & CRPS (\%) & MdAE (\%) & 90\% cov. & 90\% width (\%) & Time (s) & Epochs & $N$ \\
\midrule
Naive persistence bootstrap & 2.763 & 4.183 & 77.1 & 10.93 & 0.00 & -- & 96 \\
TEMPER & 2.824 & 3.635 & 68.8 & 6.74 & 135.10 & 264.1 & 96 \\
AR(1) bootstrap & 2.922 & 4.004 & 79.2 & 11.45 & 0.15 & -- & 96 \\
Random-walk bootstrap & 3.408 & 4.306 & 59.4 & 7.44 & 0.00 & -- & 96 \\
\bottomrule
\end{tabular}
}
\end{table}

Table~\ref{tab:benchmark} shows a close aggregate comparison between TEMPER and the naive persistence bootstrap. The naive bootstrap has the best mean CRPS, 2.763\%, while TEMPER is second at 2.824\%, a relative gap of about 2.2\%. TEMPER has the best MdAE, 3.635\%, indicating stronger median forecasts, but its 90\% intervals cover only 68.8\% of realized outcomes. The shorter interval width, 6.74\% of the origin level, confirms that TEMPER is too sharp in this configuration. The average benchmark TEMPER fit ran for 264.1 epochs under the 300-epoch cap and patience 100.

\begin{figure}[t]
\centering
\includegraphics[width=0.82\linewidth]{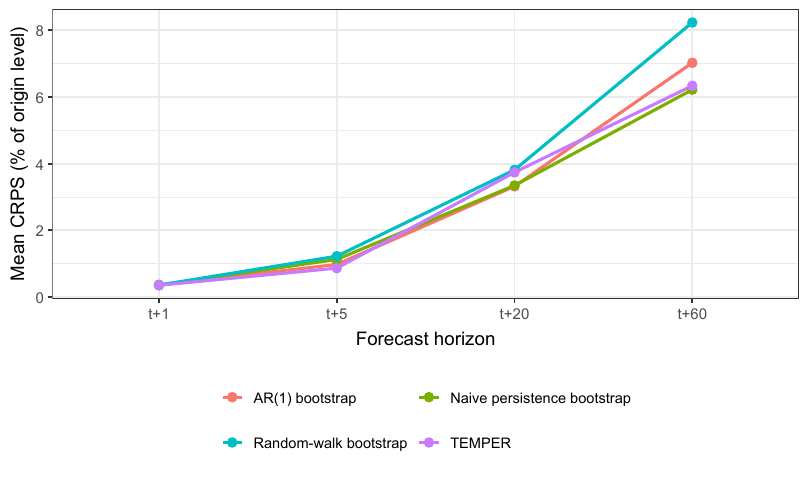}
\caption{Mean normalized CRPS by forecast horizon. TEMPER is best at $t+1$ and $t+5$, while AR(1) and naive persistence are strongest at $t+20$ and $t+60$, respectively.}
\label{fig:crps-horizon}
\end{figure}

Figure~\ref{fig:crps-horizon} decomposes the results by horizon. TEMPER has the best CRPS at $t+1$ and $t+5$, with 0.353\% and 0.867\%, respectively. AR(1) is best at $t+20$, and the naive persistence bootstrap is best at $t+60$. This pattern suggests that TEMPER learns useful short-horizon nonlinear structure in the synthetic panel, while simple horizon-wise persistence errors remain highly competitive at the longest horizon.

\begin{figure}[t]
\centering
\includegraphics[width=0.9\linewidth]{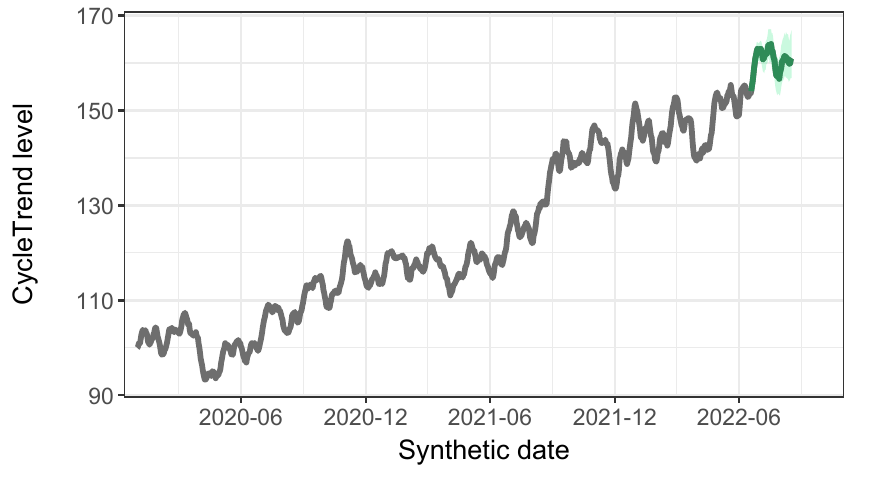}
\caption{Illustrative 60-step TEMPER fan chart for the synthetic CycleTrend series using the full available series. The chart shows the historical level path, median forecast, and central predictive interval returned by the package.}
\label{fig:fan}
\end{figure}

\begin{figure}[t]
\centering
\includegraphics[width=0.78\linewidth]{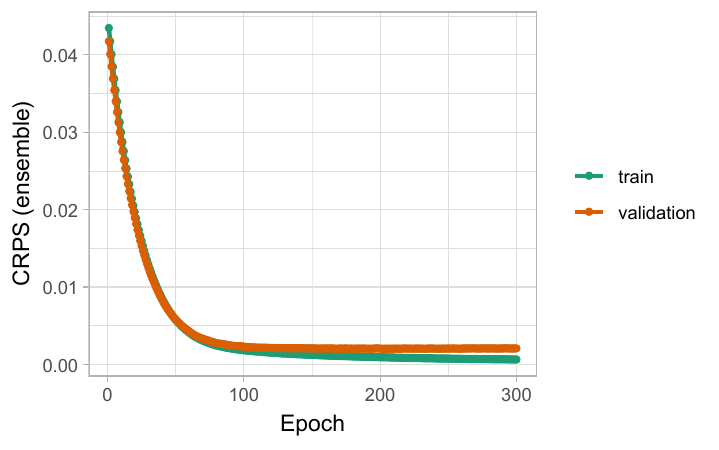}
\caption{Training and validation CRPS curves for the illustrative CycleTrend fit. The package exposes this diagnostic as part of the returned model object.}
\label{fig:loss}
\end{figure}

Figures~\ref{fig:fan} and \ref{fig:loss} illustrate the package outputs rather than the benchmark protocol. The fan chart emphasizes TEMPER's distributional interface, while the loss curve is a practical diagnostic for undertraining or overfitting.

\clearpage

\section{Paired Ablation Study}

The ablation study follows a paired design. It uses the same three synthetic series, the same four selected horizons, and the same last four rolling origins for every TEMPER variant, giving $3\times4\times4=48$ matched forecast-horizon cells per variant. Each reported CRPS delta is computed cell by cell against the full TEMPER configuration and then averaged. This avoids comparing variants on slightly different forecast tasks and makes the horizon-specific pattern visible.

\begin{table}[H]
\centering
\caption{Paired component ablation on identical series-origin-horizon cells. The paired $\Delta$ CRPS column is in percentage points of origin-level-normalized CRPS relative to Full TEMPER; negative values improve on the full model. Wins is the percentage of matched cells in which the variant has lower CRPS than Full TEMPER.}
\label{tab:ablation}
\resizebox{\linewidth}{!}{\begin{tabular}{p{0.48\linewidth}rrrrrr}
\toprule
Variant & CRPS (\%) & paired $\Delta$ CRPS & wins (\%) & 90\% cov. & 90\% width (\%) & Epochs \\
\midrule
No learned mask (near keep-all: $p_0=0.999$, $\tau=0.1$) & 2.551 & -0.472 & 39.6 & 66.7 & 7.00 & 276.6 \\
No latent bottleneck (set $d=P=60$) & 2.653 & -0.369 & 39.6 & 62.5 & 7.26 & 263.4 \\
No reconstruction (set $\lambda_{rec}=0$) & 2.674 & -0.348 & 37.5 & 64.6 & 6.89 & 270.0 \\
Reduced forest head ($M=30$, depth 4, bases 6) & 2.886 & -0.137 & 43.8 & 83.3 & 8.06 & 300.0 \\
Full TEMPER (reference: $\lambda_{rec}=0.3$, $d=10$, mask $p_0=0.8$, $M=80$, depth 5) & 3.022 & +0.000 & -- & 64.6 & 7.04 & 270.5 \\
\bottomrule
\end{tabular}
}
\end{table}

Table~\ref{tab:ablation} shows that the default full configuration is not optimal for this synthetic setting. Relaxing the learned mask gives the best average CRPS, 2.551\%, a paired improvement of 0.472 percentage points relative to the full model. Removing the latent bottleneck improves average CRPS by 0.369 points, and removing reconstruction regularization improves it by 0.348 points. The reduced forest head improves CRPS more modestly, by 0.137 points, while raising empirical coverage from 64.6\% to 83.3\%.

The win rates add a useful warning. None of the variants beats the full model on a majority of the 48 matched cells; the best win rates are 43.8\% for the reduced forest and 39.6\% for the no-mask and no-bottleneck variants. The mean improvements therefore come from large gains on some horizons rather than uniform dominance. This is exactly the kind of behavior that an aggregate-only ablation table would hide.

\begin{table}[H]
\centering
\caption{Horizon-wise paired CRPS deltas versus Full TEMPER. Entries are percentage-point changes in origin-level-normalized CRPS; negative values favor the variant.}
\label{tab:ablation-horizon}
\resizebox{0.9\linewidth}{!}{\begin{tabular}{lrrrrr}
\toprule
Variant & $t+1$ & $t+5$ & $t+20$ & $t+60$ & Mean \\
\midrule
No learned mask & +0.051 & +0.229 & -0.594 & -1.572 & -0.472 \\
No bottleneck & +0.048 & +0.358 & -0.401 & -1.482 & -0.369 \\
No reconstruction & +0.077 & +0.449 & -0.341 & -1.577 & -0.348 \\
Reduced forest & +0.006 & +0.083 & -0.858 & +0.223 & -0.137 \\
\bottomrule
\end{tabular}
}
\end{table}

\begin{figure}[t]
\centering
\includegraphics[width=0.82\linewidth]{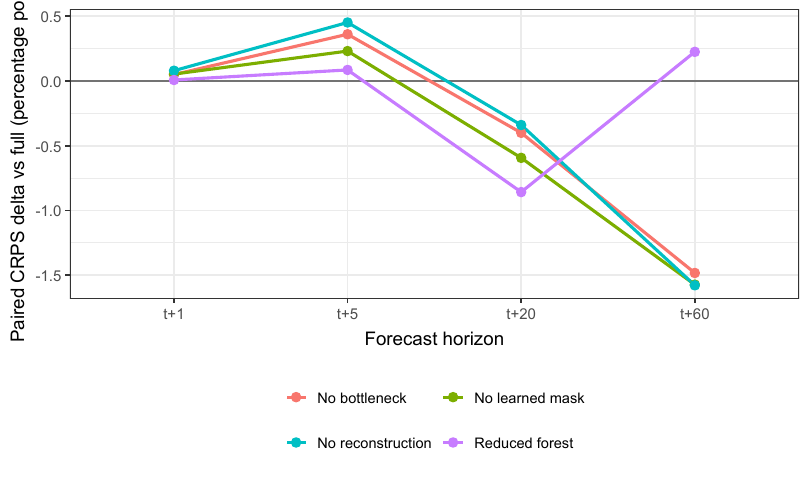}
\caption{Horizon-wise paired CRPS deltas for component variants relative to Full TEMPER. Several variants are worse at $t+1$ and $t+5$ but better at longer horizons.}
\label{fig:ablation-delta}
\end{figure}

Table~\ref{tab:ablation-horizon} and Figure~\ref{fig:ablation-delta} explain the aggregate ablation result. The variants generally degrade short-horizon CRPS but improve long-horizon CRPS. The no-mask, no-bottleneck, and no-reconstruction variants all lose at $t+1$ and $t+5$, then gain at $t+20$ and $t+60$. The reduced forest is close to neutral at $t+1$ and $t+5$, strongly better at $t+20$, but worse at $t+60$. This suggests that the full TEMPER inductive bias is useful for short horizons, whereas the mask, bottleneck, and reconstruction term may become too restrictive when the synthetic signal accumulates over longer horizons.

\section{Calibration Study}

The benchmark and paired ablation both show under-coverage. We therefore add a calibration-specific study for the full TEMPER benchmark forecasts. The first six rolling origins are used as a calibration block, and the last two origins are used as an evaluation block. All calibration rules are evaluated on the same $3\times2\times4=24$ held-out forecast-horizon cells.

Let $[\ell_i,u_i]$ be the raw 90\% TEMPER interval and $\hat{m}_i$ its median. We compare the raw interval, symmetric fixed inflation around the median by factors 1.5 and 2.0, a split-conformal residual interval using horizon-wise absolute median errors from the calibration block, and a split-conformal scale interval using horizon-wise absolute-error-to-half-width ratios. Because the calibration block is small, the split-conformal quantile is deliberately conservative. We report empirical 90\% coverage, normalized interval width, and the 90\% interval score, for which lower is better.

\begin{table}[H]
\centering
\caption{Calibration ablation on the last two rolling origins, after using the first six origins as a calibration block. IS90 is the normalized 90\% interval score; lower is better.}
\label{tab:calibration}
\resizebox{0.92\linewidth}{!}{\begin{tabular}{lrrrr}
\toprule
Calibration rule & 90\% cov. & 90\% width (\%) & IS90 (\%) & $N$ \\
\midrule
Raw TEMPER intervals & 54.2 & 6.25 & 29.47 & 24 \\
Fixed inflation (1.5x) & 79.2 & 9.37 & 19.50 & 24 \\
Fixed inflation (2.0x) & 91.7 & 12.49 & 15.07 & 24 \\
Split conformal residual & 83.3 & 21.85 & 22.24 & 24 \\
Split conformal scale & 95.8 & 17.68 & 18.41 & 24 \\
\bottomrule
\end{tabular}
}
\end{table}

Table~\ref{tab:calibration} confirms that raw TEMPER intervals are too narrow in the final evaluation block: coverage is 54.2\% with mean width 6.25\%. Fixed inflation is surprisingly effective. A 1.5-fold inflation raises coverage to 79.2\%, and a 2.0-fold inflation reaches 91.7\% coverage with the best interval score, 15.07\%. The split-conformal scale rule is more conservative, reaching 95.8\% coverage with width 17.68\% and interval score 18.41\%. The residual conformal rule reaches 83.3\% coverage but produces the widest intervals and a weaker interval score.

The calibration result is not a claim that twofold inflation is universally optimal. It shows that the current TEMPER distributional output has useful central forecasts but needs explicit calibration control. For a production or benchmark-grade version, calibration should be estimated with larger calibration windows, horizon-specific validation, and time-series-aware conformal variants.

\section{Endpoint Sensitivity}

The paired component ablation also gives a compact endpoint sensitivity analysis. It does not replace a full hyperparameter grid, but it separates the parameter axes from the component interpretation. Table~\ref{tab:sensitivity} summarizes the endpoint changes corresponding to mask relaxation, latent dimension, reconstruction weight, and forest capacity.

\begin{table}[H]
\centering
\caption{Endpoint sensitivity summary derived from the paired ablation runs. The table separates parameter axes from the component interpretation in Table~\ref{tab:ablation}.}
\label{tab:sensitivity}
\resizebox{\linewidth}{!}{\begin{tabular}{p{0.23\linewidth}p{0.38\linewidth}rrrr}
\toprule
Axis & Endpoint setting & paired $\Delta$ CRPS & wins (\%) & $\Delta$ cov. & $\Delta$ width \\
\midrule
Mask relaxation & $p_0=0.8,\tau=0.5 \rightarrow p_0=0.999,\tau=0.1$ & -0.472 & 39.6 & +2.1 & -0.04 \\
Latent dimension & $d=10 \rightarrow d=60$ & -0.369 & 39.6 & -2.1 & +0.21 \\
Reconstruction weight & $\lambda_{rec}=0.3 \rightarrow 0$ & -0.348 & 37.5 & +0.0 & -0.15 \\
Forest capacity & $M=80,D=5,K=8 \rightarrow M=30,D=4,K=6$ & -0.137 & 43.8 & +18.8 & +1.02 \\
\bottomrule
\end{tabular}
}
\end{table}

The sensitivity table supports three practical conclusions. First, the mask is the main tuning priority in this synthetic setting: moving from a stochastic learned mask toward a near keep-all mask gives the largest mean CRPS improvement. Second, the latent dimension should not be treated as a fixed compression default; the no-bottleneck endpoint performs better at longer horizons. Third, reducing forest capacity improves coverage and runtime, although the CRPS gains are smaller and horizon-dependent.

\section{Limitations}

The experiment has several limitations. First, it is synthetic, so it tests whether TEMPER can exploit controlled temporal structure rather than whether it generalizes to real operational forecasting domains. Second, the synthetic panel contains only three series and one data-generating design family. Third, the endpoint sensitivity table is not an interior hyperparameter grid. Fourth, the split-conformal calibration exercise uses few rolling origins and assumes the calibration block is informative for later origins. Fifth, the benchmark does not yet compare against standard neural forecasting systems such as DeepAR, TFT, N-BEATS, or N-HiTS on common datasets.

These limitations point to natural extensions. Better calibration-sharpness control could be obtained by horizon-wise temperature tuning, mixture regularization, conformal calibration with larger calibration blocks, or post-hoc variance scaling. The paired ablation results suggest that the learned mask, bottleneck dimension, reconstruction weight, and ensemble size should be tuned jointly and perhaps differently by horizon. Larger studies across public synthetic and real-world benchmark suites would clarify whether TEMPER's short-horizon gains persist beyond the controlled setting used here.

\section{Reproducibility}

The main experiments were run on August~31, 2026 using R 4.5.1 \cite{rcore2025}, \texttt{torch} 0.16.0 \cite{torchcran2025}, and \texttt{ggplot2} 4.0.0 \cite{wickham2016ggplot2}. The paired, sensitivity, and calibration summaries were produced from the saved raw forecast metrics on September~1, 2026. The arXiv source bundle includes the manuscript, tables, and figures needed for PDFLaTeX compilation. The ancillary directory contains the experiment script, the derived-analysis script, raw forecast metrics, summary tables, metadata, a CSV export of the generated synthetic data, and the implementation file used by the run. The exact benchmark origins, hyperparameters, generator seeds, package versions, and baseline definitions are recorded in the ancillary metadata and experiment scripts.

\section{Conclusion}

TEMPER is a compact probabilistic forecasting architecture that joins temporal encoding, differentiable masked forests, ensemble CRPS training, and mixture-based distribution reconstruction. In the synthetic rolling-origin benchmark reported here, TEMPER is competitive with the best naive baseline in aggregate CRPS, achieves the best MdAE, and has the best CRPS at $t+1$ and $t+5$. The paired ablation study gives a sharper statistical message: component effects are horizon-dependent, aggregate improvements can be driven by long-horizon gains, and calibration is a first-order issue. The paper therefore frames TEMPER as a promising and inspectable statistical machine-learning forecasting engine, with paired ablation, horizon-specific tuning, and calibrated predictive intervals as the next research priorities.

\section*{Acknowledgments}

The author thanks the R, \texttt{torch}, and PyTorch communities for the software stack used in the implementation and experiments.

\appendix

\section{Experiment Configuration}

The main benchmark uses the following TEMPER configuration: lookback 60, training forecast horizon 60, evaluated horizons $t+1$, $t+5$, $t+20$, and $t+60$, latent dimension 10, 80 soft trees, tree depth 5, initial mask keep probability 0.8, Gumbel-Softmax temperature 0.5, maximum of 8 Gaussian-mixture bases, 75\% training split, 300 maximum epochs, Adam learning rate 0.003, batch size 16, reconstruction weight 0.3, and early-stopping patience 100. The benchmark TEMPER fits ran for 264.1 epochs on average after early stopping.

The evaluated rolling origins are:
\begin{center}
\begin{tabular}{ll}
\toprule
Index & Date\\
\midrule
1 & 2021-08-17\\
2 & 2021-09-21\\
3 & 2021-10-26\\
4 & 2021-11-30\\
5 & 2022-01-04\\
6 & 2022-02-08\\
7 & 2022-03-15\\
8 & 2022-04-19\\
\bottomrule
\end{tabular}
\end{center}

The paired component ablation uses origins 5--8. The calibration study uses origins 1--6 as a calibration block and origins 7--8 as an evaluation block.

\section{Synthetic Generator Details}

The exact synthetic data generator is included in the ancillary experiment script. CycleTrend starts at 100 and uses $\phi=0.20$ with sinusoidal drift terms at periods 20 and 60. RegimeCycle starts at 80 and uses $\phi=0.15$ with alternating drift regimes every 150 observations, a 45-observation cycle, and alternating volatility regimes every 120 observations. ThresholdWave starts at 120 and uses $\phi=-0.10$ with a thresholded 36-observation sinusoid, a 90-observation cosine term, and a lagged-return correction. All series use fixed random seeds and positive level reconstruction through exponentiated clipped log returns.

\section{Implementation Notes}

The reference implementation exposes \texttt{temper()} as an R function. Its returned list contains a loss plot, a list of horizon-wise predictive functions, a fan chart, and a wall-clock training log. The predictive function list contains \texttt{dfun}, \texttt{pfun}, \texttt{qfun}, and \texttt{rfun} entries for density evaluation, cumulative probabilities, quantiles, and simulation.

\end{document}